\documentclass[letterpaper, 10 pt, conference]{ieeeconf}  

\usepackage{amsmath,amssymb,epsfig}
\usepackage{setspace}
\usepackage[final]{pdfpages}
\usepackage{cancel}
\usepackage[absolute]{textpos}
\usepackage{tikz}
\usetikzlibrary{shapes,arrows,petri,topaths}
\usepackage{tikz,fullpage}
\usepackage{tkz-berge}
\usepackage{subcaption} 
\usetikzlibrary{positioning}
\usepackage{enumerate}
\usepackage{mdframed}
\usepackage{float}
\usepackage{hyperref}	
\restylefloat{table}
\usepackage[ruled,linesnumbered]{algorithm2e}
\usepackage{bm}
\usepackage{nomencl}
\makenomenclature

\def\E{{\mathbb E}}

\IEEEoverridecommandlockouts                              

\usepackage{multicol,lipsum}

\title{\LARGE \bf
An Extended Consideration of Joint Exploration and Tracking: JET* 
}

\author{Alexander Ivanov$^{1}$ and Mark Campbell$^{2}$
\thanks{ }
\thanks{$^{1}$information 
        {\tt\small aii4@cornell.edu}}%
\thanks{$^{2}$information
        {\tt\small mc288@cornell.edu }}%
}

\newcommand{\beq}{\begin{equation}}
\newcommand{\eeq}{\end{equation}}
\newcommand{\beqa}{\begin{eqnarray}}
\newcommand{\eeqa}{\end{eqnarray}}
\newcommand{\beqan}{\begin{eqnarray*}}
\newcommand{\eeqan}{\end{eqnarray*}}

\newcommand{\R}{{\mathbb R}}

\newcounter{l1}
\newcounter{l2}
\newcounter{l3}
\newcommand{\bdotlist}{\begin{list}{$\bullet$}{}}
\newcommand{\bboxlist}{\begin{list}{$\Box$}{}}
\newcommand{\bbboxlist}{\begin{list}{\raisebox{.005in}{{\tiny
$\blacksquare$ \ \ }}}{}}
\newcommand{\bdashlist}{\begin{list}{$-$}{} }
\newcommand{\blist}{\begin{list}{}{} }
\newcommand{\barablist}{\begin{list}{\arabic{l1}}{\usecounter{l1}}}
\newcommand{\balphlist}{\begin{list}{(\alph{l2})}{\usecounter{l2}}}
\newcommand{\bAlphlist}{\begin{list}{\Alph{l2}.}{\usecounter{l2}}}
\newcommand{\bdiamlist}{\begin{list}{$\diamond$}{}}
\newcommand{\bromalist}{\begin{list}{(\roman{l3})}{\usecounter{l3}}}

\newcommand{\prop}[1]{\begin{proposition} #1 \end{proposition}}

\newcommand{\cor}[1]{\begin{corollary}   #1  \end{corollary}}

\renewcommand{\theequation}{\arabic{equation}}
\newtheorem{theorem}{Theorem}[section]
\newtheorem{exercise}[theorem]{Exercise}
\newtheorem{lemma}[theorem]{Lemma}
\newtheorem{proposition}[theorem]{Proposition}
\newtheorem{corollary}[theorem]{Corollary}
\newtheorem{definition}[theorem]{Definition}
\newtheorem{remark}[theorem]{Remark}
\newtheorem{example}[theorem]{Example}

\begin{document}

\maketitle
\thispagestyle{empty}
\pagestyle{empty}

\begin{abstract}
 Autonomous exploration and multi-object tracking by a team of agents have traditionally been considered as two separate, yet related, problems which are usually solved in two phases: an exploration phase then a tracking phase. The exploration problem is usually viewed through an information theoretic framework where a robotic agent attempts to gather as much information about the environment or an Object of Interest (OI). Conversely, the tracking problem attempts to maintain precise location information about an OI over time. This work proposes a single framework which enables the multi-robot multi-object problem to be solved simultaneously. A hierarchical architecture is used to coordinate robotic agents in the tracking of multiple OIs while simultaneously allowing the task to remain computationally efficient. The primary contributions of this work are a probabilistic constraint on the tracked OIs' covariances guarantees tracking performance throughout the entire mission. The automatic discovery of new OIs, a seamless transition to guaranteed tracking of discovered OIs, and the automatic balancing of exploration with the requirements of tracking.

\end{abstract}

\nomenclature[3]{$n_x,n_a$}{Dimention of the robot and object state}
\nomenclature[3]{$n_z,n_u$}{Dimention of measurement, and control}
\nomenclature[3]{$n_s, n_d$}{Number of GM components (mixands), for sensor and OIs}
\nomenclature[3]{$n,m \in \mathbb{N}$}{Number of robots and OIs}
\nomenclature[1]{$x_i(t), w_i(t) \in \R^{n_x}$}{Robot state and process noise}
\nomenclature[1]{$a_j(t), w_j(t) \in \R^{n_a}$}{Object state and process noise}
\nomenclature[1]{$z_{i,j}(t), v_{i,j}(t) \in \R^{n_z}$}{Measurement of object $j$ by robot $i$ and measurement noise}
\nomenclature[2]{$O_{i,j}^k \in \{0,1\}$}{Detection/observation of object $j$ by robot $i$ at time $k$}

\printnomenclature[85pt]

\section{INTRODUCTION}
The problems of exploration and tracking are tightly coupled in many real world scenarios, including surveillance, Search and Rescue (SAR), and defense. In some SAR tasks, robots first need to locate potential victims and then subsequently track them if they are moving; examples include victims in burning forests, in the ocean, or in an alpine avalanche. Furthermore, the tracking task - that of maintaining object locations - typically takes precedence over the exploration task. While exploration and tracking have typically been solved in separate stages to simplify the joint problem's complexity, this decoupled approach requires ad-hoc switching between stages which in turn makes guaranteeing tracking performance difficult. This work seeks to solve the joint problem of exploration and tracking under a unified framework, while guaranteeing tracking performance.    

As background, research has been dedicated to the exploration problem with both single robotic agents as well as groups of homogeneous and heterogeneous agents; see \cite{Robin2016} and the references within. Exploration applied to SAR includes ‘probabilistic search’, but does not usually consider tracking; several recent surveys have explored the state of this literature \cite{Robin2016,Chung2011}. This problem can also be framed as a mapping problem, as in \cite{Ivanov2016}. Of particular relevance to this work, Al Khawaldah \textit{et al.} consider the multi-robot exploration problem, but do not consider the detection or tracking of Objects of Interest (OIs) \cite{Alkhawaldah2014}. Similarly, Mottaghi and Vaughan use a particle filter to inform a team of robots of how to maximize the probability of detecting an OI \cite{Mottaghi2007}, but also do not consider tracking.

A relevant variant of the exploration problem is termed the \textit{coverage} problem, which implies ensuring that the largest possible area is sensed or `covered'. Most work on coverage is inapplicable but, Pimenta \textit{et al.} generate a continuous time algorithm which seeks to guarantee both coverage and tracking \cite{Pimenta2010}. Although \cite{Pimenta2010} attempts to solve these problems simultaneously, no guarantees of tracking performance are given, no information theoretic sense of exploration is used, and the system is fully deterministic. Elston \textit{et.al} also consider a multi-layered joint coverage and tracking problem \cite{Elston2008a}. Their work focuses on teams of `mother ships' and `daughter ships' which utilize a heuristic to partition a search area. Exploratory information is encoded by a heuristic, and robotic agents switch between pure tracking and pure coverage tasks.

A second related research area is the tracking problem from the viewpoint of traditional tracking metrics. Of particular interest,  Ferrari \textit{et al.}  consider the tracking problem in a geometric manner and develop a closed form solution to the probability of detecting OIs under linearity and observability assumptions \cite{Wei2015}. How \textit{et al.} develop an RRT based planner which seeks to maximize track detections while achieving a goal \cite{Levine2013}. Multi-agent multi-object problems have also considered questions such as data association and track assignment \cite{Robin2016}, but, in general, the tracking problem has been considered independent of the exploration or track detection problem.

Crucially, the literature rarely considers the joint exploration and planning problem. Instead, the joint problem is usually solved in two stages which are assumed independent (i.e. first explore then track).  This two-stage approach has several drawbacks. First, tracking accuracy may not be well maintained in the search phase, and OI tracks may be lost when the tracking phase begins. In many problems, robotic agents do not definitively finish the search task and continually find more OIs over time. This then necessitates switching between phases in a potentially ad-hoc way. Finally, the two stage approach fails to exploit the full ability of robot agents because agents may have excess control with which to continue to search for more OIs (victims) while maintaining track accuracy of already located OIs. 

In this work, the joint exploration and tracking (JET) problem is presented under a probabilistic framework which enables the problem to be solved in a single stage. This allows agents to utilize their full control authority, while maintaining tracking accuracy, and seamlessly transitioning between exploration and tracking. In addition, chance constraints are utilized to guarantee tracking performance, as in many applications such as SAR, tracking is paramount. An exploration objective function is derived which generalizes others found in the literature, e.g. \cite{Wei2015}. Finally, a hierarchical formulation is presented which enables the continuous optimal control problem to scale well, be solved efficiently, and maintain guaranteed tracking performance. Simulation results show the efficacy of the JET approach.

\section{Problem Formulation}
The JET problem is posed as an optimization using an information metric and probabilistic constraints. Because robotic agents must first locate OIs, the performance metric, $J_{\mathrm{info}}$, must incorporate detection up to a horizon time $T$. In applications such as SAR, tracking of \textit{all} detected OIs must be maintained, and is therefore posed as a constraint. 

The system adheres to dynamic equations:
\begin{equation}\label{eq:Dynamics}
\begin{array}{rclc} 
\dot{x}_i(t) & = & f_i(x_i(t), u_i(t), \omega_i(t)) & i \in \{1,..,n\} \\
\dot{a}_j(t) & = & F_j a_j(t) + \omega_j(t) & j \in \mathcal{T} \bigcup \mathcal{\hat{T}} \\
z_{i,j}(t) & =  & h_{i,j}(x_i(t), a_j(t), \nu_{i,j}(t)) & u_i(t) \in \mathcal{U}_i\\
\end{array}
\end{equation}

\noindent
where $f_i(\centerdot)$ and $F_j(\centerdot)$ are the dynamics of robotic agents and OIs respectively. Note that LTI dynamics and Gaussian noise are assumed only for OIs. The sets $\mathcal{T}$ and $\hat{\mathcal{T}}$ denote tracked and untracked OIs respectively. It is assumed that each robot can localize itself via a nonlinear Kalman Filter (KF), and each robot's state estimate, $\bar{x}_i(t)$, and covariance are available \cite{Bar-Shalom2001, Thrun2005}. The precise type of KF is inconsequential to the results in this work. Discrete-time measurements are utilized. Thus, $z_{i,j}$ returns a value only at discrete-time instances. Note: $t^k = k \Delta t$.

\subsection{Objective Function}
Because the framework of the problem requires agents to detect un-tracked OIs, an analysis of detection probability is in order. The probability of a single OI being detected at time $k$ is given as:

\begin{equation} \label{eq:ProbDetect_raw}
 P(\mathcal{O}^k_j=1|\bm{z}_j^{1:k-1}, \bm{u}^{1:k-1})
\end{equation}

\noindent
where $(\mathcal{O}^k=1)$ denotes a positive detection at time instant $k$, $\bm{z}_j^{1:k-1}$ is a vector of all sensor measurements taken of the particular un-tracked OI up until the current time instance $k-1$, and $\bm{u}^{1:k-1}$ is the sequence of controls given to the robotic agents up until time $k-1$ (a zero-order-hold assumption). The \textbf{bold} notation implies that this is an aggregated vector of all similar variables, i.e. $\bm{u}$ is an aggregate of the controls of all robotic agents. This expression can be decomposed using the law of total probability un-marginalize the robotic and un-tracked object states. Then, by sequentially conditioning the detection likelihood, robotic state, and object state on all other variables, Eq. \eqref{eq:ProbDetect_raw} can be shown equivalent to: 

\begin{multline} \label{eq:intProbDetect}
\int \limits_{\bm{x}^k \in \bm{\mathcal{X}}^k} \int \limits_{a^k_\mathrm{j} \in \mathcal{A}^k_j} \bigg( p(\mathcal{O}^k_j=1| \bm{x}^k,  a^k_{j})\times \\ 
p(\bm{x}^k| \bm{u}^{1:k-1}) p(a^k_{j}|\bm{z}_j^{1:k-1}, \bm{u}^{1:k-1}) \bigg) d\bm{x}^k da^k_j
\end{multline}

\noindent
where, $\bm{x}^k$ is the state of the robotic agents at time $k$, and $a^k_{j}$ is the state of the $j$th OI. Notice that the first term of \eqref{eq:intProbDetect} is the detector model while the second and third terms are the predictive agent and OI distributions respectively. The full derivation of \eqref{eq:intProbDetect} is shown in Appendix \ref{appendix:Deriv_Prob_Detect}.
	
To understand the detection probability - Eq. \eqref{eq:intProbDetect}- more easily, consider that there is only one agent and its position is perfectly known (i.e. the integral with respect to $x^k$ disappears). In addition, suppose there exists a perfect OI detector with a circular field of view of radius $r$. In this case, Eq. \eqref{eq:intProbDetect} reduces to:

\begin{multline} \label{eq:intProbDetectSimplified}
\int\limits_{(a^k_{j}| d(a^k_{j}, {x}^k) \leq r)} p(a^k_{j}|{z}_j^{1:k-1}, {u}^{1:k-1}) da^k_{j}
\end{multline}

\noindent
In this case, the probability of detection simplifies to be the probability that the OI is within sensor range of the agent. The detection probability in \eqref{eq:intProbDetectSimplified} corresponds to, and therefore \eqref{eq:intProbDetect} generalizes, the sensor function in \cite{Wei2015}. 

Using \eqref{eq:intProbDetect} as an objective function maximizes the probability that a single OI is detected at the next time instant $k$. Let $\bm{x}(t)$ and $\bm{a}(t)$ be the states of all robotic agents and OIs at time $t$. The variable  $\bm{z}^{1:k-1}_j$ is a vector of sensor measurements taken of object $j$ up to time $t^{k-1}$,  and $\bm{u}(t)$ are the controls given to the robotic agents. The variables $ \bm{\omega}(t)$ and $\bm{\nu}(t)$ are stochastic noise affecting the state of the mobile objects (robots and OIs), and measurements respectively. In this scenario, $ J_{\mathrm{info}}$ is  written as:
\begin{equation}\label{eq:Obj_Func_Specific}
\begin{array}{ll} 
J_{\mathrm{info}}(\bm{x}, \bm{a},\bm{z}, \bm{u}, \bm{\omega}, \bm{\nu}, k) := & \\  \hspace{20pt} \sum\limits_{j \in \mathcal{\hat{T}}} P(\mathcal{O}^{k:K}_j=1|\bm{z}^{1:k-1}_j, \bm{u}^{1:k-1})
\end{array}
\end{equation}

\noindent
Note that the dependence of the arguments on time is suppressed for compactness and denoted by the argument $k$. The summation is taken over the set of un-tracked OIs $\mathcal{\bar{T}}$. The total number of tracked and un-tracked OIs, $m$, is unknown, finite, and assumed to be $m \leq n$. The challenge that $m$  is unknown is addressed when discussing the hierarchical approximation of this problem in section \ref{sect:ApxObj}. 

\subsection{Constraint on Tracking Performance}
In traditional tracking, objects are typically tracked using a selection from a set of standard tracking estimators, referred to as Kalman Filtering (KF) techniques, which includes variants for both linear and non-linear dynamics \cite{Bar-Shalom2001}. In this work, tracked OI states are estimated using a KF, which models the state transition and measurement likelihoods as Gaussian distributions. As such, tracking performance can be analyzed via the covariance matrix:

\begin{equation}
\Sigma^k_{j} = \E[(\bar{a}^k_{j}- a^k_{j})'(\bar{a}^k_{j}- a^k_{j})]  \quad \forall j \in \mathcal{T}
\end{equation}

\noindent
where $\bar{a}^k_j$ is the estimate of the $j$th tracked OI at $k$. 

To guarantee tracking of discovered OIs, a bound on the covariance, $ \Sigma^k_{j}$, must be satisfied for every tracked OI. Since all real-world sensors are imperfect, there is a chance that the sensor does not detect/measure a tracked OI at a particular time instance, i.e. missed or intermittent detection. The presence of intermittent measurements implies that no deterministic bound can be given for $\Sigma^k_{j}$; instead, a bound on the expected value is used, i.e. $\E \big[ \Sigma^k_{j} \big] :=\int p(\Sigma^k_j|\mathbf{Z}_j^{1:k})p(\mathbf{Z}_j^{1:k})dZ$.

For cases of linear dynamics and intermittent measurements, the Algebraic Riccatti Equation is a contraction and can be used to guarantee the existence of $\E \big[ \Sigma^k_{j} \big]$  \cite{Zhou2016}, \cite{Censi2011} ,\cite{Plarre2009}. Through these results, it can be shown that, for a certain range of probabilities of detection $P(\mathcal{O}_j^k = 1)$, the tracking error covariance $ \E \big[ \Sigma^k_{j} \big]$ is bounded uniformly in $t$ and there exists a finite steady state distribution for $\Sigma^k_{j}$  \cite{Zhou2016, Plarre2009}. Thus, one only needs to bound the probability of detecting an OI in order to provide a bound on its expected covariance matrix. 

Consider now the following bound on the tracking error of tracked OIs, 

\begin{equation} \label{eq:ProbDetect_Constr}
P(\mathcal{O}^k_{j}=1|\bm{z}_j^{1:k-1}, \bm{u}^{1:k-1}) \geq 1 - \alpha \quad \forall j \in \mathcal{T}
\end{equation}

\noindent
where $\alpha \in (0,1)$ is a user set parameter. Notice that this bound utilizes the same expression as the objective function in Eq. \eqref{eq:ProbDetect_raw}. The constraint in \eqref{eq:ProbDetect_Constr} simply says that the joint robotic system has at least a probability of $1-\alpha$ of seeing OI $j$ up to the discrete time instant $k$, given the measurement and control histories. Assuming that OIs follow LTI dynamics with additive Gaussian noise, the results in \cite{Zhou2016, Plarre2009} can be leveraged and $ \E \big[ \Sigma^k_{a^k_j} \big]$ can be computed for the chosen $\alpha$. This calculation can be done via direct Monte-Carlo simulation is in the results here, or by using the bound derived in \cite{Zhou2016}. The proposed optimal control problem is then: maximize \eqref{eq:Obj_Func_Specific} subject to (\ref{eq:Dynamics},\ref{eq:ProbDetect_Constr}). This formulation implies that robots are opportunistic explorers, but must maintain tracking. 

\section{A Hierarchical Approximation}
The problem presented in Section II is intractable, and the only known general way to solve this problem is Dyanamic Programming (DP). In addition, the problem does not fulfill the standard DP assumption of an additive, time-invariant reward $J_\mathrm{info}$ . Finally, the dimensionality of the control and state spaces increases linearly with number of robotic agents and OIs, and trajectory optimization scales poorly with state dimension \cite{Bertsekas2005a} \cite{Betts2009}.

To make the JET problem tractable, a hierarchical approximation is presented. The hierarchical framework first provides optimal Next-Best-View (NBV) positional goals at horizon time $T$ to each robotic agent. A lower level nonlinear optimization then solves the continuous time optimal control problem for each agent independently and satisfies the dynamics. This approach sacrifices information optimality by only coordinating the terminal location of robots, but still solves the exploration and tracking problem jointly and provides probabilistic tracking guarantees.

\subsection{Reduction to the NBV problem} \label{sect:ReductionNVB}
The NBV optimization is designed to give fast approximate exploration goals, at the time horizon $T$, to each robotic agent. These goals seek to maintain tracking performance. Three assumptions reduce the problem in Section II to the NBV problem. First, note the objective in Eq. \eqref{eq:Obj_Func_Specific} is dependent on the trajectory history $(\bm{x}_{0:T}, \bm{a}_{0:T},\bm{z}_{0:T})$. This motivates the use of an Open-Loop-Feedback (OLF) strategy in which expected intermediary measurements are ignored \cite{Bertsekas2005a}, therefore we assume no measurements are taken between the initial time $t^0$ and the terminal time $T$. Second, to make \eqref{eq:Obj_Func_Specific} additive, we assume no `information overlap' occurs between robotic agents when the agents are 'well spaced';  i.e. if $||x_i^K-x_j^K||$ is large enough then $ P(\mathcal{O}^K_j=1|\bm{z}_j^{1:k-1}, \bm{u}^{1:k-1}) \approx \sum_{i=1}^n P(\mathcal{O}^K_{i,j}=1|\bm{z}_{j}^{1:k-1}, \bm{u}_i^{1:k-1})$. Last, a coarse, discrete, linear approximation of the robotic dynamics is assumed. Formally: $\exists E\subset \R^{n_x}, \mathrm{U} \subset \R^{n_u}, B \in \R^{n_x \times n_u} \quad s.t. \quad E = \{x: x = B \mathrm{u}, \mathrm{u} \in \mathrm{U}\}$, and $\forall x \in E$, $x$ is reachable by the nonlinear system from the origin in time $T$. This is a local controllability assumption. Work has been done on approximating local reachability \cite{Bayen2007}. In this work, an optimal linear approximation is not derived, but, for the unicycle model used here, a simple analysis can yield a coarse approximation readily \cite{Thrun2005}. In the sequel, note that $K \Delta t = T$. Finally, to help satisfy constraint \eqref{eq:ProbDetect_Constr} at $K$, linear observability at $K$ through $H_{i,j} \in \R^{n_x \times n_z}$ and a linearized transition $F_i \in \R^{n_x \times n_x}$ are assumed. Thus, the coarse dynamics and measurement predict motion up to the time horizon $T$:
\begin{equation}\label{eq:Coarse_Dynamics}
\begin{array}{rcl} 
x_i^K & \approx & F_i  x_i^0 + B_i  \mathrm{u}^0_i + \mathrm{w}^0_i \\
a_j^K & = & F_j^K \cdot a_j^0 + \mathrm{w}_j^0 \\
z_{i,j}^K & =  & H_{i,j}(x_i^K- a_j^K)\cdot(x_i^K- a_j^K) +  \mathrm{v}_{i,j}^0\\
\mathrm{u}_i^0 & \in & \mathrm{U}_i
\end{array}
\end{equation}

\noindent
The variables $(\mathrm{w}, \mathrm{v})$ are the time integrals of their continuous time noise counterparts \cite{Bar-Shalom2001}. Roman notation, (e.g. $u$ vs $\mathrm{u}$), denotes a discrete time counterpart of a variable. Because this coarse approximation follows approximate linear dynamics, the control sets $\mathcal{U}_i$ and $\mathrm{U}_i$ are not the same.  Note that the OIs are assumed to follow LTI dynamics, which allows for the application of bounded expected covariance due to \eqref{eq:ProbDetect_Constr}.  

\subsection{An approximate objective function for NBV goals} \label{sect:ApxObj}
 Given the coarse dynamics and measurement prediction in \eqref{eq:Coarse_Dynamics}, the higher level exploration problem seeks to maximize:

\begin{multline} \label{eq:Apx_Obj_Highlevel}
\max_{\bm{\mathrm{u}}^0\in \bm{\mathrm{U}}} \hat{J}_{\mathrm{info}}(\bm{x}^0, \bm{a}^0,\bm{z}^0, \bm{\mathrm{u}}^0, \bm{\mathrm{w}}^0, \bm{\mathrm{v}}^0, t^0) = \\
\max_{\bm{\mathrm{u}}^0\in \bm{\mathrm{U}}} \sum_{j\in \mathcal{\hat{T}}} P(\mathcal{O}^K_{j}=1|\bm{z}^{0}, \bm{\mathrm{u}}^{0})
\end{multline}
Equation \eqref{eq:Apx_Obj_Highlevel} maximizes detection of untracked OIs at the horizon time $K$. More specifically, the optimal control, $(\bm{\mathrm{u}}^{0})^*$, produced by maximizing Eq. \eqref{eq:Apx_Obj_Highlevel}, generates an optimal set of next-best viewpoints $(\bm{x}^{K})^*$.

\subsection{A tighter tracking constraint}
 The NBV formulation, while at a coarser level,  must continue to guarantee tracking for each OI in $\mathcal{T}$.  Instead of satisfying Eq. \eqref{eq:ProbDetect_Constr}, the NVB formulation requires that a robotic agent is assigned to each OI which is actively being tracked. The assigned agent is then required to guarantee an observation of its OI at the horizon time $T$. This constraint is formally defined as:
\begin{equation}
\begin{array}{rcl} \label{eq:ProbDetect_Constr_Assigned}
\exists i \in \mathcal{A}, \hspace{20pt} \forall j \in \mathcal{T} \hspace{8pt}  s.t. \\  P(\mathcal{O}^K_{i,j}=1|\bm{z}^{1:k-1}_{j}, \mathrm{u}^{1:k-1}_i) & \geq & 1 - \alpha
\end{array}
\end{equation}
where $\mathcal{A}$ is the set of assigned robots. Equation \eqref{eq:ProbDetect_Constr_Assigned} is a tighter constraint than Eq. \eqref{eq:ProbDetect_Constr}, and the assignment of agents necessitates the assumption that $m \leq n$. The higher level NVB problem is summarized as: maximize Eq. \eqref{eq:Apx_Obj_Highlevel} subject to Eqs. (\ref{eq:Coarse_Dynamics},\ref{eq:ProbDetect_Constr_Assigned}). The NVB result is a set of optimal Next-Best-View points $(\bm{x}^K)^*=\bm{x}^*(T)$. As $T \rightarrow \Delta t$, the NVB problem guarantees a probability of detecting known OIs at each time step, but greatly reduces the exploratory capability of agents, and makes exploration myopic. If robotic agents have overlapping sensor fields of view at time $T$, the mutual information between agents must be considered. This problem is related to Distributed Data Fusion and can be a difficult to solve \cite{Campbell2016}. Instead, via assumption two, an additional `well spaced' constraint separates agents by $M$, a positive sensor-dependent constant, at the terminal time:
\begin{equation} \label{eq:ExplorationSeparation}
||\bar{x}_i^K -\bar{x}_j^K|| \geq M \quad \forall i \neq j \quad i,j \notin \mathcal{A}
\end{equation}

\subsection{Distributed optimization}
Given the maximizer of the higher level problem, $(\bm{x}^{K})^*=\bm{x}^*(T)$, which implicitly assigns the robotic agents to OIs, the low level problem is considered. The constraint in Eq. \eqref{eq:ProbDetect_Constr_Assigned}, along with the results in \cite{Zhou2016,Censi2011,Plarre2009}, ensures that tracking performance is maintained for a short time horizon $T$. Thus, the lower level problem no longer needs to consider tracking performance directly. Instead, the following constraint must be met:

\begin{equation} \label{eq:Terminal_State_Constr}
\bm{x}(T) = \bm{x}^*(T)
\end{equation}

\noindent
The lower level optimization is then solved independently at each $t^k$, in a distributed fashion, by each robot. A single assumption, consistent with the OLF strategy, is needed to make the low level problem tractable. Recall that $J_{\mathrm{info}}$ is non-additive through time since there is `information overlap' between a robot's sensor through time. This also means that $J_{\mathrm{info}}$ is time varying and also dependent on the previous path taken by robotic agents. The only way to accurately account for this time and state dependence would be to enlarge the state space of the optimization to include the time varying state of the cost function \cite{Bertsekas2005a}. Although this is theoretically possible, the subsequent state space explosion makes state augmentation impractical. For the lower level problem, we instead take the line integral of the current probability of detection over the planned robotic path. In other words:
\begin{equation} \label{eq:CostLL}
	J_{\mathrm{LL}} := \int\limits_{x_i(t^k:T)}   P(\mathcal{O}^k_j=1|\bm{z}_j^{1:k-1}, \bm{u}_i^{t^k:t}) dt
\end{equation}

\noindent
This is in accordance with assumption one in Section \ref{sect:ReductionNVB}. Intuitively, Eq. \eqref{eq:CostLL} uses assumption one but positions the robot such that if a measurement is unexpectedly taken at any time $t \in (t^k,T)$, the robot will be in a locally optimal position. The lower-level optimization is now given by:

\begin{equation} \label{eq:Objective_LowLevel}
\begin{array}{llcl}
\displaystyle
\max_{u_{i}(t)} & J_{\mathrm{LL}}(x_i, \mathbf{a},z_i, u_i, \omega_i, \bm{\nu}, t^k) & & \\
\textrm{s.t.} & \dot{x}_i(t)  =  f_i(x_i(t), u_i(t), \omega_i(t)) &  & \\
&\dot{a}_j(t)  =  F_j a_j(t) + \omega_j(t) & & \\
&z_{i,j}(t)  =   h_{i,j}(x_i(t), a_j(t), \nu_{i,j}(t)) & & \\
&x_i(T) = x_i^*(T) & & \\
&u_i(t)  \in  \mathcal{U}_i, \quad t \in (t^k, T), \quad j \in  \mathcal{\hat{T}} \cup \mathcal{T}
\end{array}
\end{equation}

\noindent
The lower level optimization is solved independently at each $t^k$, in a distributed fashion, by each robot. At each time step $k$, new information is incorporated in the posterior distribution of unknown object locations and the robots re-optimize \eqref{eq:Objective_LowLevel}.

\subsection{The Joint Exploration and Tracking (JET) algorithm}
A full description of a Joint Exploration and Tracking (JET) algorithm with probabilistic guarantees can now be given in Alg. \eqref{alg:JET}. 

\begin{algorithm} 
	$\mathbf{x}(0) = $ InitializeStates()\;
	$p(\bm{a}) = $ InitalizeOiDist() \;
	$\mathcal{T} = \emptyset$ \;
	\While{true}{
		$UpdateOiPdf(\mathcal{I}(t), \bm{z}(t))$ \label{alg:Update}\;
		\If{ \textrm{New OI detected} \label{alg:NewDetection}}
			{$\mathcal{T} = \{ \max{(\mathcal{T})}+1 \}\cup \mathcal{T}$}
		\If{(t\%T)==0) or New OI detected}
		{
			\If{(t\%T)==0}{$t_0 = t;$}
			$Assignment = SolveAssignment(\mathcal{I}(t))$ \label{alg:Assignment} \;
			$(\bm{x}^*(T+t_0)) = SolveNBV(\mathcal{I}(t)); \label{alg:NBV}$
		}

		\For{$i \in \{1,...,n\} $}{
			$(x_i^*(t), u^*_i(t)) = SolvePath(\mathcal{I}(t))$ \label{alg:ContinousOPT}\;
			$\mathrm{u}^*_i = u^*_i(t+\Delta t/2);$
		}
		$t = t+\Delta t$ \;
	}
	\caption{Joint Exploration and Tracking: JET} \label{alg:JET}
\end{algorithm}

The JET algorithm is structured intuitively. For compactness, all currently available information is denoted $\mathcal{I}(t)$, including estimates of robot states, OI estimates and distributions, and previous controls. At each discrete time instant $k$, any new sensor measurements are used to update the OI distributions (line \ref{alg:Update}). If a new OI is detected (line \ref{alg:NewDetection}), it is added to the detected set $\mathcal{T}$. Using the new detected set, the assignment problem is solved so that each detected OI is assigned a robotic agent, in order to satisfy Eq. \eqref{eq:ProbDetect_Constr_Assigned} (line \ref{alg:Assignment}). In this study, the assignment is performed by solving the linear assignment problem using Euclidean distance from agents to expected OI locations at $T$ as the cost. Different metrics for solving the assignment problem are possible and would result in switching behavior being manifest under different conditions \cite{Elston2008a}.  

Given the assignment, the high level, Next-Best-View, problem is then solved, Eqs. (\ref{eq:Coarse_Dynamics} - \ref{eq:ExplorationSeparation}) (line \ref{alg:NBV}), using current estimates of the OIs expected positions. The NVBs , $(\bm{x}^K)^*$, for each robotic agent are then used to solve the decentralized lower level path planning problem locally using direct transcription to account for the non-linear dynamics (line \ref{alg:ContinousOPT}) \cite{Betts2009}. Note that the high level NVB problem is solved centrally at each time horizon, or when new OIs are discovered; each robotic agent can solve its own optimization independently in parallel.

\section{PERFORMANCE, MODELING, AND JET GUARANTEES} \label{sect:Modeling}
Given the defined higher and lower level optimization problems outlined in Alg. \eqref{alg:JET}, the specific modeling assumptions on the distributions of untracked and tracked OIs, robots' states, and the form of the detection function are required to enable full implementation. 

First, consider the detection function in Eq. \eqref{eq:intProbDetect}. Most current literature assumes a perfect detector, which is an indicator function within a detection range or region \cite{Wei2015}. Real sensors are seldom perfect detectors, and their accuracy may drop off as a function of range. This decrease in accuracy is especially true when using sensors, such as LIDAR and cameras, which become less accurate or resolute at a further range.

Instead of an indicator function, this work models the idealized detector as an un-normalized Gaussian Mixture (GM), which has several advantages. First, the detection probability can be cast as a function of distance, which is appropriate for LIDAR and camera sensors. Second, GMs can be composed to model complex behaviors. Finally, GMs have continuous derivatives. 

Formally, the state space of the OIs and agents are assumed the same, i.e. $n_a=n_x$, and the detector for tracked and un-tracked OIs  is modeled as a quasi-concave GM where $c_l$ are constant vectors:
\begin{multline} \label{eq:GM_Sensor}
P(\mathcal{O}^k=1| x^k,  a^k) =  \hspace{80pt} \zeta_l > 0\\ \sum_{l=1}^{n_s} \zeta_l \cdot \exp \bigg(-\dfrac{1}{2}(a^k-(x^k-c_l))'\Sigma_{O_l}^{-1}(a^k-(x^k-c_l))\bigg) \\
\mathrm{where}: \quad \max(p(\mathcal{O}^k=1| x^k,  a^k=x^k)) =1
\end{multline}

KF methods are used to estimate tracked OI and robot states yielding multivariate Gaussian distributions.
\begin{equation} \label{eq:Robot_KnownTarget_Dist}
\begin{array}{rcll}
p(x_i^k| u_i^{1:k-1}) & \sim & \mathcal{N}\big( \bar{x}_i^k, \Sigma_i^k \big) \\
p(a_j^k|z^{1:k-1}, \bm{u}^{1:k-1}) & \sim & \mathcal{N}\big( \bar{a}_j^k, \Sigma^k_{j} \big), & j \in {\mathcal{T}} 
\end{array}
\end{equation}

Finally, versatility and smoothness make GMs a good candidate to represent the untracked OI distribution as well. Recent powerful tools which have been developed to approximate arbitrary distributions as GMs, \cite{Bishop06}, are of particular importance to the objective function in Eq. \eqref{eq:Obj_Func_Specific} because, if a robotic agent fails to detect an untracked OI at any time instant, a `negative' measurement usually results in a posterior distribution which has no closed form. Therefore after every negative measurement, the resultant posterior distribution must be re-approximated as a GM using K-means clustering or another technique in order to maintain computational efficiency and avoid having to discretize the exploration space. As a result, unknown OIs are modeled using GMs:

\begin{multline} \label{eq:GM_Unknown}
p(a^k_{\mathrm{j}}|z^{1:k-1}, \bm{u}^{1:k-1}) = \hspace{40pt} j \in \hat{\mathcal{T}}, \gamma_l>0 \\ \sum_{l=1}^{n_d} \gamma_l \cdot \exp \bigg(-\dfrac{1}{2}(a^k_{jl}-\mu_{jl})'\Sigma_{jl}^{-1}(a^k_{jl}-\mu_{jl})\bigg) 
\end{multline}

Note that if tracked OIs are estimated using a Kalman filter, this implies that, as soon as a new OI is discovered and enters the tracked set $\mathcal{T}$, the tracked OI's distribution is no longer modeled as a the GM in Eq. \eqref{eq:GM_Unknown}. Therefore, \textit{only} negative measurements are taken of the un-tracked OIs. 

Given these modeling assumptions, solution characteristics emerge, which are instrumental in allowing the lower level path planning problem to be solved quickly using standard non-linear optimization tools.

\prop{\label{prop:CVX_Constraint} Given Eqs. (\ref{eq:GM_Sensor}, \ref{eq:Robot_KnownTarget_Dist}), if $n_s=1$, then the feasible set for the assigned robot $i$ in Eq. \eqref{eq:ProbDetect_Constr_Assigned} contains a subset $\mathcal{S}$ which is convex in $\bar{x}_i(T)$ and quadratic in $||\bar{x}_i(T) - \bar{a}_j(T)||$}

The proof of Proposition (4.1) is found in Appendix \ref{appendix:Proof_Of_Prop}. The convexity of the constraint provides a fast global feasibility check and is therefore valuable in making the JET algorithm real-time. For details on the difficulty of finding initial feasible points see Phase I methods in \cite{boyd2004convex}. A similar, but weaker, result is true if the sensor is modeled by more than one mixand, assuming the detector GM is quasi-concave. This result allows for a much larger class of sensors to be accurately modeled while maintaining the convexity of the NVB problem. Both the proof and further discussion are in Appendix \ref{appendix:Proof_Of_Prop}.

\begin{figure*}[t]
	\centering
	\subcaptionbox{Excess Authority: $0.3$m/sec\label{fig:SingleRobV1}}[.31\linewidth][c]{%
		\includegraphics[width=.31\linewidth]{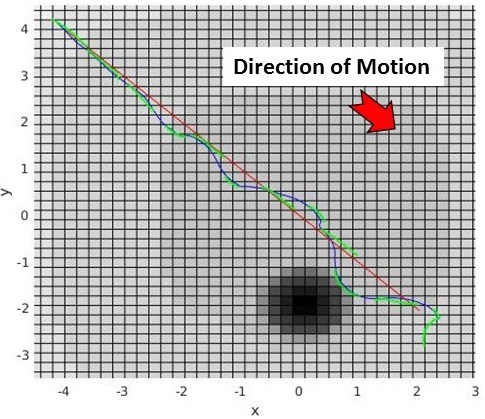}}\quad
	\subcaptionbox{Excess Authority: $0.8$m/sec\label{fig:SingleRobV25}}[.31\linewidth][c]{%
		\includegraphics[width=.30\linewidth]{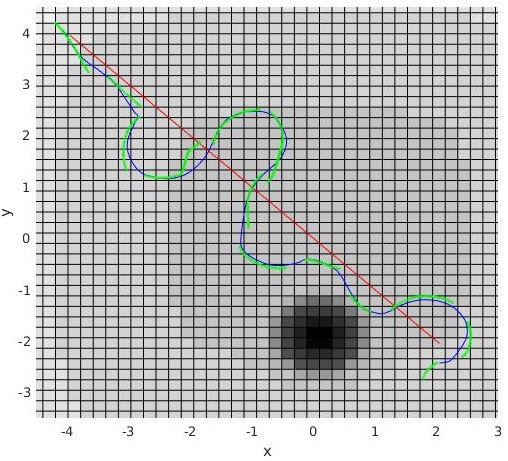}}\quad
	\subcaptionbox{Excess Authority: $2.3$m/sec\label{fig:SingleRobV30}}[.31\linewidth][c]{%
		\includegraphics[width=.31\linewidth]{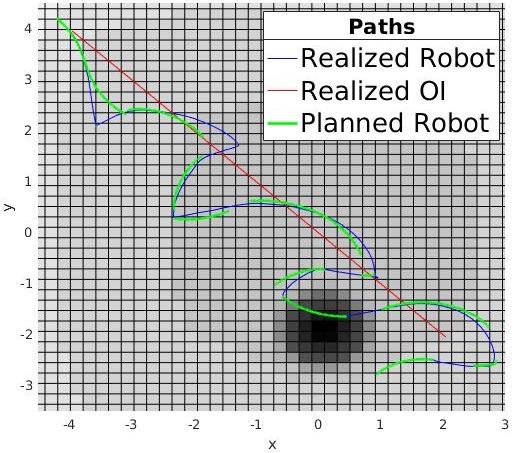}}
	\caption{Robot behavior as a function of control authority.}\quad
	\label{fig:SingleRobot}
\end{figure*}

\section{SIMULATION RESULTS} 

To analyze the behavior of the JET algorithm, three simulation studies are conducted. The first study shows the behavior of the low-level optimizer as a function of the excess control authority given to a robotic agent. The second study analyzes the behavior of the high-level optimizer and its ability to distribute robotic agents and maintain tracking of OIs. The final study analyzes the behavior of the OI covariance matrix under the JET algorithm's assumptions.

\subsection{Exploration as a function of control authority}
In this case study, a single robot is tasked with maintaining track accuracy of a single OI which has already been detected. Furthermore, there is an \textit{a priori} probability of another OI being in the space, which is nearly uniform except for a peak at the bottom right corner of the map. This study seeks to exemplify the behavior of the low-level planning component of the algorithm as additional control authority is made available. In addition, the study analyzes the effects of the Open Loop Feedback (OLF) approximation on the optimality of the resultant path. 

Recall that an OLF strategy does not take into account any expected measurements of the environment in the future; as such, it is sub-optimal in general. Instead, with each new measurement of the map, the robot re-plans its path to incorporate the new information. Thus, the repeated re-planning addresses the sub-optimality of the OLF assumption. In this case study, the robotic agent moves deterministically and assumes process noise in the tracking model of the OI. The true behavior of the OI is deterministic to ensure repeatability of the experiment. In all cases, the following variables are kept constant: the prior distribution on untracked OIs, the realized tracked OI motion, and the initial conditions. The only variable is the control authority, $u_i(t)$, available to the robotic agent. 

Figure \ref{fig:SingleRobot} shows the results of this case study for three levels of control authority. The blue line is the robot's true realized path throughout the experiment, while the red line is the path taken by the tracked OI. The green lines in Fig. \ref{fig:SingleRobot} show planned paths at particular instances in time separated by $0.5$sec. Measurements of the environment are taken at $10Hz$ and the initial prior distribution on untracked OIs is overlayed as a heat map, with darker colors implying higher probability. Notice that, in each of the figures, the robot strays away from the expected OI path to explore the surrounding area. In all cases, the robot attempts to sense the \textit{a priori} peak in the bottom right and stays to the left of the true object trajectory. As the control authority increases ($a \rightarrow c$), the robot is able to explore further from the expected OI trajectory, which it must observe at intervals of $2.0$sec. Subsequently, the planned green paths are  longer. The planned and realized (blue) paths differ due to new measurements, which update the heat-map (not shown), but are very similar as measured in euclidean distance. This suggests that the OLF method approximates the optimal trajectory well even while ignoring future measurements.

\subsection{Behavior and speed of the hierarchical approach}
The second study seeks to qualitatively assess the ability of the JET algorithm to utilize a team of robots to both explore for unknown OIs and maintain localization of tracked OIs. The time horizon is set at $T =2$sec and the detection/observation constraint probability for tracked OIs is set at $.65$ ($\alpha = .45$).

\begin{figure}
	\centering
	\subcaptionbox{\label{fig:MultiRobT0}$t \in [0 \rightarrow 1.5]$}[.93\linewidth][c]{%
		\includegraphics[width=.75\linewidth]{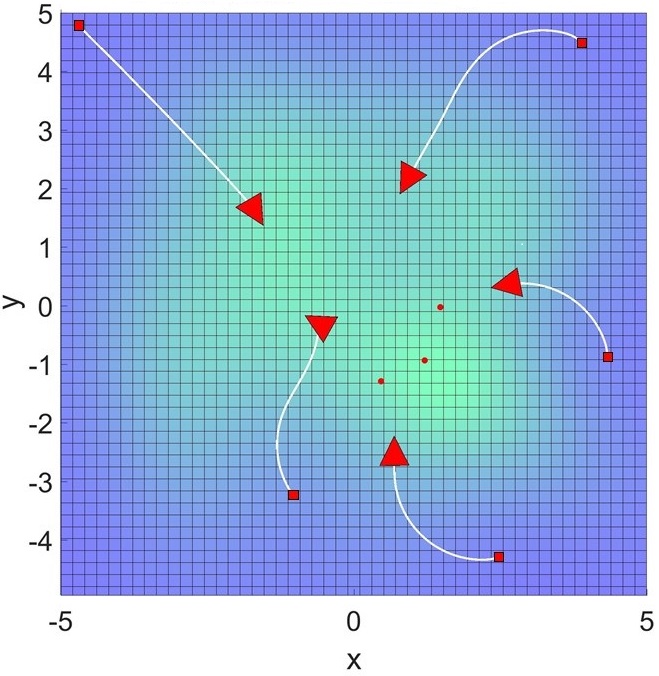}} 
	\subcaptionbox{\label{fig:MultiRobT2}$t \in [2 \rightarrow 3.5]$}[.93\linewidth][c]{%
		\includegraphics[width=.75\linewidth]{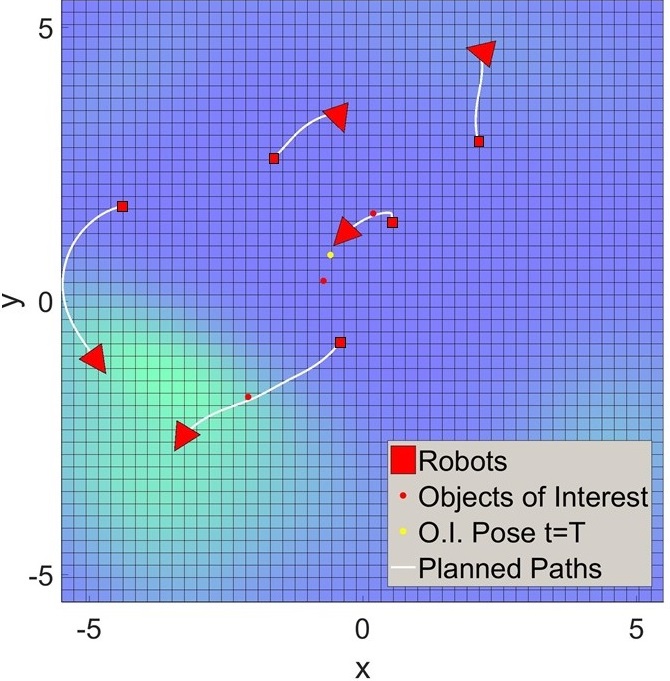}}
	\subcaptionbox{\label{fig:MultiRobT8}$t \in [8 \rightarrow 10.5]$}[.93\linewidth][c]{%
		\includegraphics[width=.75\linewidth]{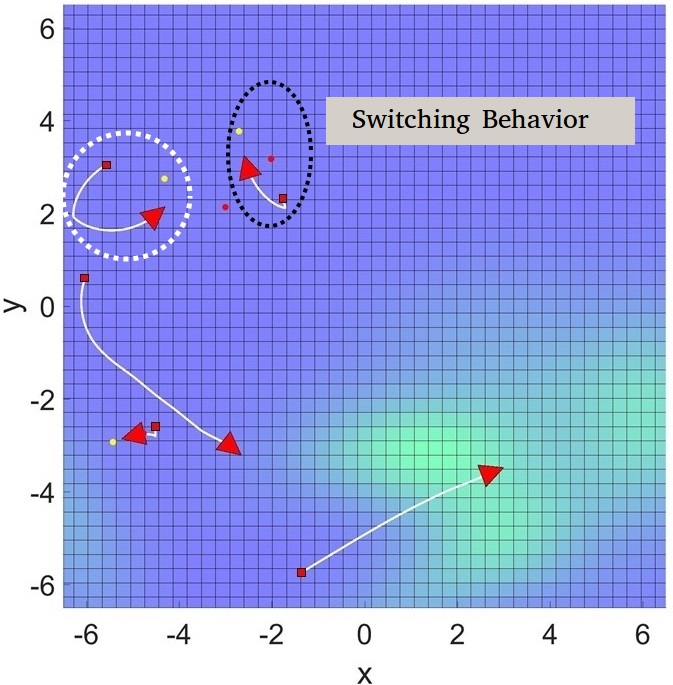}}
	\caption{Simulation studying multi-agent behavior, with the transition from exploration to tracking. The predicted path is shown up to $T-.5$ to help visually distinguish expected OI positions}
	\label{fig:MultiRob}
\end{figure}

To avoid clutter and confusion, Fig. \ref{fig:MultiRob} shows a series of snapshots of planned agent paths at three successive times $(t \in \{0,2,8\})$. Each figure shows five robotic agents searching a $12\times12$ area for, and subsequently tracking, three unknown OIs. The OIs move with stochastic dynamics. Initially, no information is known about the number or location of OIs except that they are not near the current location of the robotic agents. As such, the same un-informative prior distribution is assumed for the location of each OI's initial location. Since un-tracked OIs are independent and have not yet been detected by definition, their distribution develops in \textit{exactly the same way} due to the negative measurements taken by robotic agents. The recursively updated posterior distribution of all un-tracked OIs is represented as a GM distribution, as shown by the heat-maps in Figure \ref{fig:MultiRob}, where warmer colors represent a greater likelihood of an un-tracked OI being located at that point. 

Agents are shown in red, with their initial position marked by a square and their planned final position and orientation shown as a triangle. True OI positions are shown as red dots, while estimated positions at $T$ are shown as yellow dots. Figure \ref{fig:MultiRobT0} shows the initial condition $(t = 0\mathrm{sec})$ and all untracked OIs have no expected position. Figure \ref{fig:MultiRobT2} $(t = 2\mathrm{sec})$ shows a single OI being tracked, while Fig. \ref{fig:MultiRobT8} $(t = 8\mathrm{sec})$ shows all OIs being tracked.

The high level NBV optimization places the terminal positions of exploratory agents near the peaks of the unknown objects' GM distribution (light green) while keeping the robots' fields of view well separated. Conversely, at the time horizon $T$, assigned robots are required to be relatively near the expected terminal positions of their assigned OIs in order to satisfy the detection constraint. The low-level continuous time path planner then optimizes the robots' trajectories, thereby improving information gathering, while satisfying the robots' non-linear dynamics. Each figure shows that the terminal locations of purely exploratory robotic agents are well separated due to the separation constraint in Eq. \eqref{eq:ExplorationSeparation}. As a consequence, the planned exploratory trajectories tend to also be separated without requiring explicit coordination between robots at the low level. 

A closer examination of the individual robotic paths (not shown) reveals a similar behavior to that seen in Fig. \ref{fig:SingleRobot}, where the robotic agent explores its environment when it has excess control authority, and tracks its assigned OI. The robotic agent's path crosses the OI's expected path approximately at each time horizon $T$. Figure \ref{fig:MultiRobT8} $(t = 8\mathrm{sec})$ shows that after six time horizons, the robotic team successfully discovers all OIs and maintains tracking of each OI. 

A new emergent `switching' behavior from the JET algorithm can be seen in Fig. \ref{fig:MultiRobT8}. A robot, near ($-2,2$), which is tracking an OI near the robot's current position, changes its role to track a newly discovered OI, near ($-2,3$). The robot and its new OI are encircled with dashed black lines. At the same time, a robot that was previously exploring is now tasked with tracking the `old' OI (encircled with dashed white lines). Similar switching occurs when new OIs are discovered, and exploratory vs tracking roles are changed/updated. 

\subsection{Distribution of tracked OI covariance}

\begin{figure}
	\includegraphics[width= 1 \linewidth]{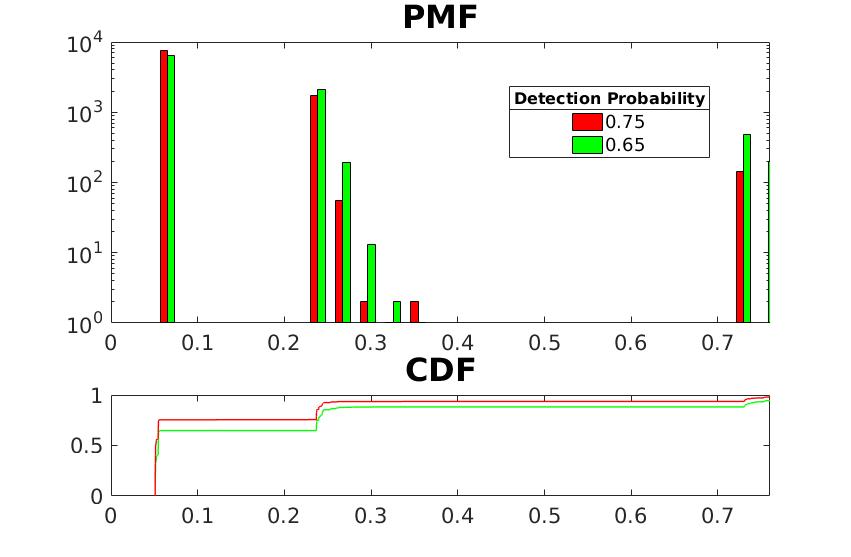}
	\caption{The empirical PMF and CDF of $\Sigma$ norms at $t = 20T$.}
	\label{fig:SigmaSS}		
\end{figure}

In \cite{Censi2011}, the author discusses the discrete behavior of the distribution of covariance matrices in LTI systems with intermittent measurements. The work here uses an LTI model of OI behavior, which is noise driven in the velocity states, and therefore meets the assumptions in \cite{Zhou2016,Censi2011}. Figure \ref{fig:SigmaSS} shows a histogram of 10,000 Monte-Carlo, covariance norms of a single tracked OI after 20 time horizons. The OI is intermittently measured at each time horizon. This is equivalent to assuming that the coarse-dynamics tracking constraint in \eqref{eq:ProbDetect_Constr_Assigned} is met for all 20 time horizons.

In this study, a probabilities of detection of $0.65$ and $0.75$ are analyzed. In addition, the same time horizon of $T=2$sec, and OI process and measurement noise statistics are used as compared to the prior studies. Figure \ref{fig:SigmaSS} shows only the smallest $95\%$ of samples to avoid poor figure scaling. Notice that peak of the PMF, representing 64.8\% and 75.4\% of all cases respectively, are clustered at $< 0.1\mathrm{m}^2$; this implies that the OI is still being well tracked. The means of the PMFs are $0.265\mathrm{m}^2$ ($1-\alpha = .65$) and $0.154 \mathrm{m}^2$ ($1-\alpha = .75$). The CDFs show that $87.7\%$ and $93.5\%$ of covariance matrices respectively, have a norm less than $.27 \mathrm{m}^2$. A norm of $.27\mathrm{m}^2$ indicates that the OI has a $99.7\%$ chance of being within two sensor radii of its mean position. Summarizing all cases, if the probabilistic constraint of detection in Eq. \eqref{eq:ProbDetect_Constr_Assigned} is met for each time horizon, a robotic agent maintains relatively accurate positional awareness of an OI in $87.7\%$ and $93.5\%$ respectively. This style of Monte-Carlo study can be used to inform a designer of the appropriate trade in time horizon length, tracking accuracy, and exploratory performance for a particular application of the JET algorithm. 

\section{Practical and Real Time Implications}

The likelihood that OI tracking is lost if the constraint in Eq. \eqref{eq:ProbDetect_Constr_Assigned} is met is small, yet the approximate hierarchical JET algorithm may lose tracking of an OI more frequently than the constraint implies. If the covariance of tracked OIs becomes large enough, a single robot may not be able to guarantee viewing its assigned OI with the required $(1-\alpha)$ probability. In this case, the problem becomes infeasible. If a tracked OI's covariance grows too large, a hybrid behavior is necessary to make the high level problem feasible once more. This hybrid behavior may require a tracking robot to switch to a pure tracking/following criterion and ignore the cost function in Eq. \eqref{eq:Objective_LowLevel} completely until tracking accuracy is recovered. Fortunately, the convexity of the constraint in Eq. \eqref{eq:ProbDetect_Constr_Assigned} allows for a fast feasibility check.

In addition, the lower level optimization is subject to limitations experienced by Non-Linear Program (NLP) solvers \cite{Betts2009}. This work uses the MIT DRAKE toolbox along with a direct transcription method to solve the continuous time optimal control problem \cite{Tedrake2016}. The initial guess provided to the NLP solver can make a large difference in computation time and even feasibility. In addition, the existence of a non-trivial gradient of the objective function in Eq. \eqref{eq:Obj_Func_Specific} is required for a timely solution. Although not common, a locally near-zero gradient may occur when a robotic agent is much faster than its assigned OI. 
 
The average computational performance was evaluated on a 2.6Ghz Intel i7-6600U processor in MATLAB using the DRAKE toolbox and SNOPT as the underlying NLP solver. The combined time of high level problem with 5 agents and a single robotic agent's low level problem typically falls in a range of $(.5-6\mathrm{sec})$. This excludes instances in which the NLP solver has difficulty converging due to a near-zero gradient, which can be compensated for by using appropriate problem scaling. In summary, these initial results show that the JET algorithm has potential as a real-time planner.

\section{CONCLUSIONS}

This work proposes a framework which enables the multi-robot multi-object problem to be solved simultaneously. The JET algorithm allows a team of robots to search for OIs while a probabilistic constraint on the tracked OIs' covariances guarantees tracking performance throughout the entire mission. Automatic discovery of new OIs, a seamless transition to guaranteed tracking of discovered OIs, and automatic balancing of exploration with the requirements of tracking are the primary novelties of the proposed algorithm. A novel hierarchical architecture is used to approximate the optimal control problem and coordinate robotic agents in the tracking of multiple OIs while simultaneously allowing the task to remain computationally efficient. 

The JET algorithm enables each robotic agent to fully utilize its control authority to optimize exploratory behavior. At the same time, JET provides probabilistic guarantees on tracking performance, which are crucial in search and rescue scenarios. Simulation results show that the JET algorithm produces intuitive exploratory paths while maintaining tracking accuracy. In addition, the JET algorithm is able solve for continuous time optimal trajectories in a receding horizon fashion and has real-time performance potential.



\bibliographystyle{ieeetr}
\bibliography{Heirarchicla_Exp_Conf}

\section*{APPENDIX}

This Appendix provides derivations and proofs for some of the formulas discussed in the paper. 

\subsection{Derivation of probability of detection}
\label{appendix:Deriv_Prob_Detect}

A derivation of the probability of detecting a single OI after a single time-step is given. In the following derivation, the measurement subscript $j$ is dropped; measurements of other objects have no effect on the $j$th object since objects are assumed independent.

\begin{equation} \label{eq:probDetect}
P(\mathcal{O}^k_j=1|\bm{z}^{1:k-1}, \bm{u}^{1:k-1})
\end{equation}

Using the law of total probability we obtain 

\begin{equation}\label{eq:probDetectBayes}
\dfrac{P(\mathcal{O}^k_j=1,\bm{z}^{1:k-1}, \bm{u}^{1:k-1})}{p(\bm{z}^{1:k-1}, \bm{u}^{1:k-1})}
\end{equation}

Focusing just the numerator, we can un-marginalize the location of the robot and the location of the untracked OI.

\begin{multline} \label{eq:ProbDerivLine1}
\int \limits_{\bm{x}^k \in \mathcal{X}^k} \int \limits_{a^k_\mathrm{j} \in \mathcal{A}^k_j} p(\mathcal{O}^k_j=1,\bm{z}^{1:k-1}, \bm{u}^{1:k-1}, x^k,  a^k_j) d\bm{x}^k da^k_j = \\
\int \limits_{\bm{x}^k \in \mathcal{X}^k} \int \limits_{a^k_\mathrm{j} \in \mathcal{A}^k_j} p(\mathcal{O}^k_j=1|\bm{z}^{1:k-1}, \bm{u}^{1:k-1}, \bm{x}^k,  a^k_j)* \\ p(\bm{z}^{1:k-1}, \bm{u}^{1:k-1}, \bm{x}^k,  a^k_j) d\bm{x}^k da^k_j 
\end{multline}

The first multiplicative likelihood, $p(\mathcal{O}^k_j=1|\bm{z}^{1:k-1}, \bm{u}^{1:k-1}, \bm{x}^k,  a^k_j)$, term is independent of previous object observations and previous control inputs. Thus Eq. \eqref{eq:ProbDerivLine1} is equivalent to:

\begin{multline} \label{eq:ProbDerivLine2}
\int \limits_{\bm{x}^k \in \mathcal{X}^k} \int \limits_{a^k_\mathrm{j} \in \mathcal{A}^k_j} p(\mathcal{O}_j=1| \bm{x}^k,  a^k_j)* \\ p(\bm{z}^{1:k-1}, \bm{u}^{1:k-1}, \bm{x}^k,  a^k_j) d\bm{x}^k da^k_j = \\
\int \limits_{\bm{x}^k \in \mathcal{X}^k} \int \limits_{a^k_\mathrm{j} \in \mathcal{A}^k_j} p(\mathcal{O}_j=1| \bm{x}^k,  a^k_j) p(\bm{x}^k| \bm{u}^{1:k-1})* \\ p(\bm{z}^{1:k-1}, \bm{u}^{1:k-1}, a^k_j) d\bm{x}^k da^k_j
\end{multline}

Notice that the second multiplicative likelihood term, $p(\bm{x}^k| \bm{u}^{1:k-1})$ , is the predictive robot distribution. Continuing, \eqref{eq:ProbDerivLine2} is equivalent to:

\begin{multline}
\int \limits_{\bm{x}^k \in \mathcal{X}^k} \int \limits_{a^k_\mathrm{j} \in \mathcal{A}^k_j} p(\mathcal{O}_j=1| \bm{x}^k,  a^k_j) p(\bm{x}^k| \bm{u}^{1:k-1})* \\  p(a^k_j|\bm{z}^{1:k-1}, \bm{u}^{1:k-1}) p(\bm{z}^{1:k-1}, \bm{u}^{1:k-1})d\bm{x}^k da^k_j
\end{multline}

Noticing that the final multiplicative term is independent of the integration, and that it cancels with the denominator in Eq. \eqref{eq:probDetectBayes}, Eq. \eqref{eq:intProbDetect} gives the result \QED.

\subsection{Proof of Prop. 4.1 and constraint properties} \label{appendix:Proof_Of_Prop}

\proof{ Given Eqs. \eqref{eq:GM_Sensor}, \ref{eq:Robot_KnownTarget_Dist} an expression in terms of $\bar{x}_i^k$ for the following set is desired:
	
\begin{equation} \label{eq:Contr_Set}
\{\bar{x}^k_i|P(\mathcal{O}^K_{i,j}=1|z^{1:k-1}_{j}, \mathrm{u}_i^{1:k-1}) \geq 1- \alpha \} 
\end{equation}

Using Eqs. \eqref{eq:intProbDetect},\eqref{eq:GM_Sensor},\eqref{eq:Robot_KnownTarget_Dist} the LHS of the condition in \eqref{eq:Contr_Set} is a function of $\bar{x}^k$, and is proportional to a GM distribution. Since the tracked object is represented by a single Gaussian, the expression reduces to: 

$$
\Bigg\{\bar{x}_i^k \Big|  \sum_{l=1}^{n_s} \zeta_l \cdot  \mathcal{N}_{\bar{x}_i^k} \big( c_l + \bar{a}_j^k , \Sigma_{O_l} + \Sigma_i^k + \Sigma_j^k \big) \geq 1- \alpha \Bigg\} 
$$

By taking a log, and using Jensen's inequality the following condition produces an inner bound for \eqref{eq:Contr_Set}:

\begin{multline} \label{eq:Jensens_Constr}
\sum_{l=1}^{n_s} \ln\Big(\frac{\zeta_l} {|2 \pi \big( \Sigma_{O_l} + \Sigma_{x_i^k} + \Sigma_{a_j^k} \big)|^{1/2}} \Big) - \\ \frac{1}{2}*(\bar{x}_i^k - c_l - \bar{a}_j^k)'\big( \Sigma_{O_l} + \Sigma_i^k + \Sigma_j^k \big)^{-1} (\bar{x}_i^k - c_l - \bar{a}_j^k) \\ \geq  \ln(1- \alpha) 
\end{multline}

Let $\tilde{\Sigma} = \Sigma_{O_1} + \Sigma_i^k + \Sigma_j^k $, if $n_s=1$, then:

\begin{multline} \label{eq:Quadratic_Constr}
(\bar{x}^k_i - c_1 - \bar{a}^k_j)'\tilde{\Sigma}^{-1} (\bar{x}^k_i - c_1 - \bar{a}^k_j) \leq \\  -2\ln \Bigg(\frac{(1- \alpha) |2 \pi \tilde{\Sigma}|^{1/2}}{\zeta_1}\Bigg)
\end{multline}

\noindent
Thus $\mathcal{S}$ is defined by Eq. \eqref{eq:Quadratic_Constr}, a convex quadratic in $\bar{x}_i^k - \bar{a}_j^k$ \QED.}

Notice that if the right hand side of \eqref{eq:Quadratic_Constr} is negative, the tracking constraint cannot be satisfied and the problem is infeasible. 

Since $c_l$ have so far been arbitrary in the more general case of \eqref{eq:Jensens_Constr}, there is no formula for the set which satisfies the inequality  in terms of $\bar{x}^k$. Sufficient properties for the general GM sensor modal are now shown.

\cor{ Suppose the sensor is modeled as a \textit{quasi-concave} GM, and that tracked and agent distributions are Gaussian. If there exists some $\bar{x}^*$ which maximizes Eq. \eqref{eq:ProbDetect_Constr_Assigned} and is feasible for $\alpha =0$, then the set defined by Eq. \eqref{eq:ProbDetect_Constr_Assigned} has a subset $\mathcal{S}$ which is convex and non-empty for $\alpha \in [0,\alpha_{\mathrm{max}}]$, $\alpha_{\mathrm{max}} < 1$. \label{corr:CVX_Constraint}}

\textit{Proof:} Corollary \eqref{corr:CVX_Constraint} is a direct consequence of the properties of quasi-convex functions \QED.

The constraint defined in \eqref{eq:Jensens_Constr} is an under-bound on the true constraint set and is much simpler to compute. Notice that the function the the LHS in of Eq. \eqref{eq:Jensens_Constr} has similar properties to the softmax function, where the largest argument tends to dominate. Roughly speaking, a sufficiently clustered set of mixand means $c_l$ should ensure that \eqref{eq:Jensens_Constr} has a a solution and can be used instead of Eq. \eqref{eq:ProbDetect_Constr_Assigned}.

\subsection{The assignment problem}
\begin{figure}
	\centering
	\includegraphics[scale=.9]{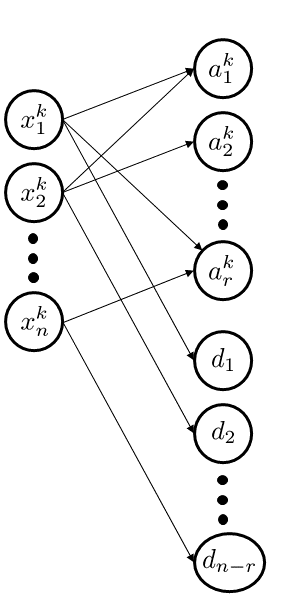} 
	\caption{The Assignment Problem. Some connections are omitted for clarity.}
	\label{fig:Assignment}
	\vspace{-10pt}
\end{figure}

In this formulation, in order to guarantee tracking performance the algorithm first assigns at least one robot to each tracked OI. To do this, a variant of the linear assignment problem can be solved. First define a mathematical graph (Fig. \ref{fig:Assignment}) with edges between robot nodes and OI nodes. In addition provide $n-r$ fictitious nodes for each of the $n-r$ robots which will remain un-assigned; here $r$ is the cardinality of $\mathcal{T}$.

The graph only connects robots to objects which can be reached within $T$. The weight of these corresponding edges is simply the euclidean distance between them. In addition, each robot is connected to the $n-r$ dummy nodes with cost zero. 

The given setup is really a simplification of the true underlying assignment problem. As it has been stated above, this problem allows for a fast solution using the Hungarian algorithm or its variants. It does not take into account information gain differences from assignment. In addition, it does not capture behaviors such as "double teaming" an OI to ensure observation. Finally, it does not take into account either robot or OI uncertainties. This may be significant when considering that an OI with low uncertainty may be un-tracked some time while uncertain OIs must be tracked immediately. At the same time, the closest robot to a well known OI may gain more information by switching OIs.



\addtolength{\textheight}{-12cm}   


\end{document}